\documentclass[letterpaper, 10 pt, conference]{ieeeconf}
\usepackage{cite}
\usepackage[dvipsnames]{xcolor}

\usepackage{times}
\usepackage{graphicx}
\usepackage{amssymb}
\usepackage{gensymb}
\usepackage{amsmath}
\usepackage{breakurl}
\usepackage{multirow}
\usepackage{adjustbox}
\usepackage{lipsum}
\usepackage{hyperref}
\usepackage{cleveref}

\let\labelindent\relax
\usepackage{enumitem}
\usepackage{url}
\usepackage{booktabs}
\usepackage[ruled,vlined,linesnumbered]{algorithm2e}
\let\svsection\section
\let\svsubsection\subsection
\let\svsubsubsection\subsubsection

\def\section{\xdef\Unit{section}\svsection}
\def\subsection{\xdef\Unit{subsection}\svsubsection}
\def\subsubsection{\xdef\Unit{}\svsubsubsection}

\let\svlabel\label
\def\label#1{\expandafter\xdef\csname Unit-#1\endcsname{\Unit}\svlabel{#1}}

\def\Unitref#1{\csname Unit-#1\endcsname~\ref{#1}}

\begin{document}
\title{Integrated Motion Planner for Real-time Aerial Videography \\with a Drone in a Dense Environment}

\author{Boseong Felipe Jeon
        and~H. Jin Kim% <-this % stops a space
\thanks{*This material is based upon work supported by the Ministry of Trade, Industry \& Energy(MOTIE, Korea) under Industrial Technology Innovation Program. No.10067206, 'Development of Disaster Response Robot System for Lifesaving and Supporting Fire Fighters at Complex Disaster Environment'}% <-this % stops a space
\thanks{Department of mechanical and aerospace engineering,
        Seoul national university of South Korea
        {\tt\small \{a4tiv,hjinkim\}@snu.ac.kr}}%
}

\newcommand{\configspace}{\mathbf{\chi}}
\newcommand{\free}{\mathbf{\mathbf{\chi}}_{free}}
\newcommand{\obs}{\mathbf{\mathbf{\chi}}_{obs}}
\newcommand{\vis}{\mathbf{\mathbf{\chi}}_{vis}}
\newcommand{\occ}{\mathbf{\mathbf{\chi}}_{occ}}
\newcommand{\dobs}{\mathbf{\mathbf{\chi}}_{dobs}}
\newcommand{\docc}{\mathbf{\mathbf{\chi}}_{docc}}
\newcommand{\state}{{{s}}}
\newcommand{\R}{{\mathbb{R}}^{3}}
\newcommand{\RS}{{\mathbb{R}}^{6}}
\newcommand{\visfield}{\psi(v;v_p)}
\newcommand{\EDT}{\mathrm{EDT}(v;\obs)}
\newcommand{\seq}{{\sigma}_{[t_0,t_N]}}
\newcommand{\transiscore}{c_v(v_{n-1},v_{n})}
\newcommand{\visscore}{\psi_v(v;\state_p)}
\urlstyle{tt}

\maketitle

\begin{abstract}
This letter suggests an integrated approach for a drone (or multirotor) to perform an autonomous videography task in a 3-D obstacle environment by following a moving object. The proposed system includes 1) a target motion prediction module which can be applied to  dense environments and 2) a hierarchical chasing planner based on a proposed metric for visibility. In the prediction module, we minimize observation error given that the target object itself does not collide with obstacles. The estimated future trajectory of target is obtained by  covariant optimization. The other module, chasing planner, is in a bi-level structure composed of \textit{preplanner} and \textit{smooth planner}. In the first phase,  we leverage a graph-search method to preplan a chasing corridor which incorporates safety and visibility of target during a time window. In the subsequent phase, we generate a smooth and dynamically feasible path within the corridor using  quadratic programming (QP). We validate our approach with multiple complex scenarios and actual experiments. The source code can be found in \url{https://github.com/icsl-Jeon/traj_gen_vis}.
\end{abstract}

% \fi
\IEEEpeerreviewmaketitle

\section{Introduction} 
\label{intro}
Video filming has been one of the most popular applications of unmanned aerial vehicles equipped with vision sensors,   utilizing their maneuverability and improvement in the technologies such visual odometry \cite{qin2018vins} and mapping \cite{hornung2013octomap,oleynikova2018safe}. For example, drones have been employed in various cinematographic tasks from personal usage to broadcasting sport events, and corresponding research has received great interests in the recent decade \cite{nageli2017real,penin2018vision,bonatti2018autonomous}. Still, the automation of the videographic tasks using drones remains as an open challenge especially in general dense environments.   

This letter addresses an online motion strategy developed for more realistic situations where multiple obstacles have arbitrary shapes and the future trajectory of target is not exactly known a priori to the filming drone except the location of sparse via-points which are pre-selected for filming purposes. Also, we do not assume that the arrival time at each point is known to the drone. For example, a drone can be deployed for the cases such as shooting a ski game or racing  where players are supposed to pass defined spots in a track. As another common example, we can consider an event where an important person (or actor) passes through defined locations in a crowded place and the arrival times for the spots are not exactly predetermined.  

\begin{figure}[!t]
\centering
\includegraphics[width=0.4\textwidth]{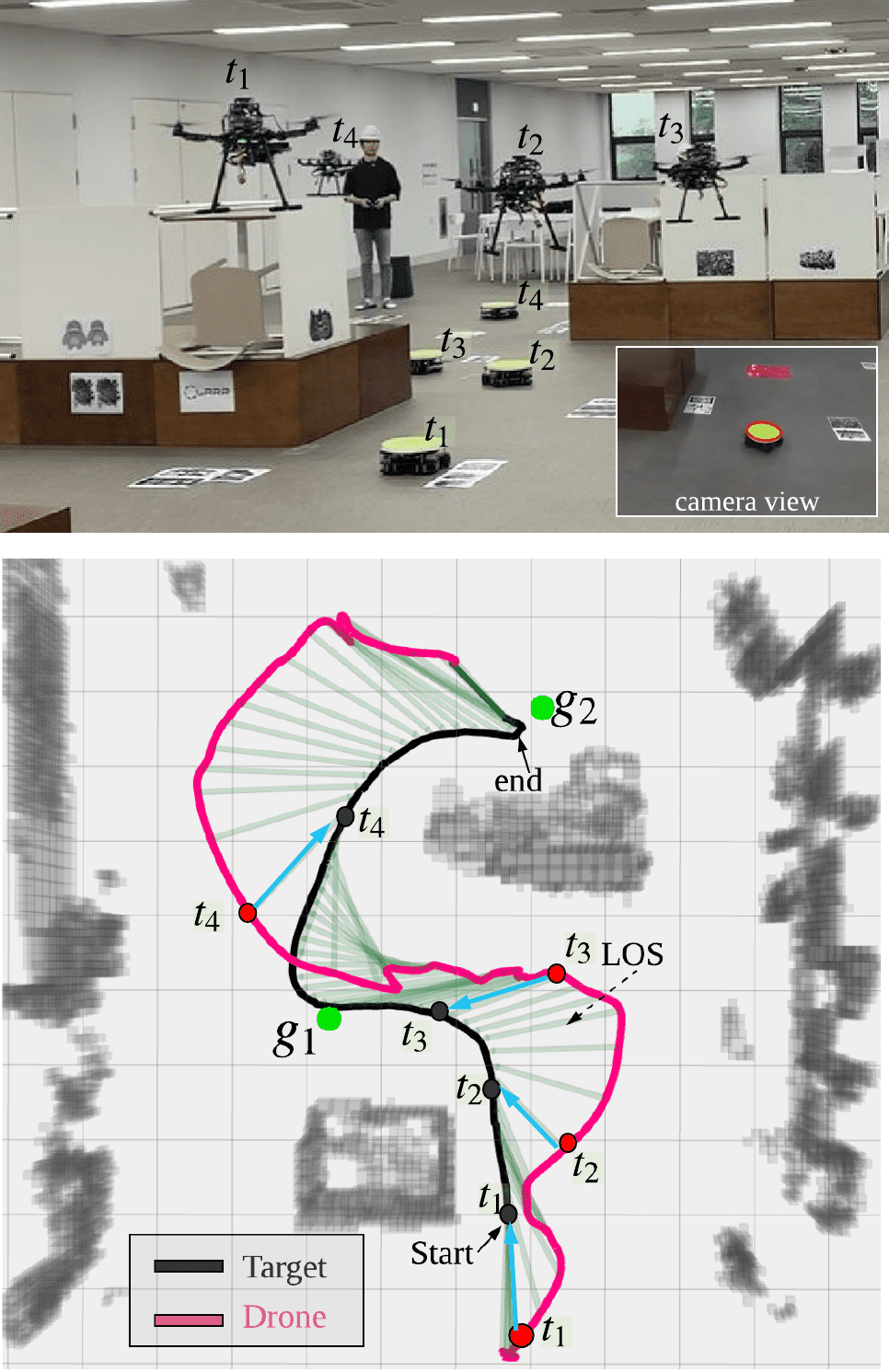}
\DeclareGraphicsExtensions.
\caption{\textbf{Top:} Autonomous aerial video shooting using a drone fora  moving target in plane classroom with cluttering objects. The drone plans a chasing trajectory on-the-fly to incorporate safety and visibility of target against obstacle. The target (UGV with a green marker) is driven manually by a human operator. \textbf{Bottom:} Path history of the target (black) and chaser(magenta). The history of line-of-sight (LOS) of the drone toward the target is visualized with sky-blue arrows.}
\label{fig_intro}
\end{figure}

\subsection{Technical challenges} 
    In our problem, the followings can be pointed out as main challenges, which should be handled jointly.
    \subsubsection{Smooth transition \label{sss:one}}
        first of all, the smoothness of flight path of a drone is essential for flight efficiency  avoiding jerky motion, which could cause increased actuation inputs and undesirable shooting quality.  
    \subsubsection{Flight safety} 
    the recording agent should be able to maintain its safety against arbitrary shape of obstacles not only simple obstacles (e.g. ellipse or sphere) for the broad applicability.  
    
    \subsubsection{Occlusion against obstacles of general shape}
    Occlusion should be carefully handled in obstacle environments. It could degrade the aesthetic quality of video, which could be one of the top priorities in cinematographic tasks. More practically, a duration of occlusion of a dynamic object might interrupt the autonomous mission if the cinematographer drone fails to re-detect the target object after losing the target out of the field of view.   
    
    \subsubsection{Trade-off between global optimality and fast computation}
    As described in 1)-3), a motion strategy for videography in the considered cases aims to achieve multiple objectives simultaneously. Such a multi-objective problem is subject to local minima and might yield a poor solution if it relies entirely on numerical optimization. On the contrary, if one relies only on sampling or discrete search algorithm such as RRT* \cite{karaman2011sampling} and A*\cite{duchovn2014path} to focus on the global optimality at the cost of online computation, a drone might not be able to respond fast enough for the uncertain motion of target on-the-fly. Therefore, trade-off between optimality and fast computation should be taken into account in a balanced manner.
    \subsubsection{Target prediction considering obstacles}
     for the operation in obstacle environments based on incomplete information of target trajectories, another challenge is a reliable motion prediction of a dynamic object with consideration of obstacles. For example, if a chaser makes infeasible prediction without properly reflecting obstacle locations, the planning based on wrong prediction will also become unreliable. Thus, accounting for obstacles is crucial to enhance the chasing performance in our scenario.   
\subsection{Related works} 

The previous works \cite{nageli2017real}, \cite{penin2018vision} and \cite{bonatti2018autonomous} addressed the similar target following problem with consideration of flight efficiency, safety and visibility (A1-A3) under continuous optimization formulation to deal with A1-A4 their problem settings. \cite{nageli2017real} and \cite{penin2018vision} developed a receding horizon motion planner to yield dynamically feasible path in real-time for dynamic situations. They assume an ellipsoidal shape for obstacles, which is not applicable to more general cases, having difficulty in fully satisfying A2 and A3. Also, the objective function contain multiple non-convex terms such as trigonometry and product of vectors. This formulation might not be able to produce a satisfactory solution in short time due to local-minima as discussed in A4.   

In \cite{bonatti2018autonomous},  occlusion and collision was handled in a general environment which is represented with octomap. Nevertheless, they relied on numerical optimization of the entire objectives  containing complex terms such as integration of signed distance field over a manifold. Such approach might not be able to guarantee the satisfactory optimality, similar to the case of \cite{nageli2017real,penin2017vision} (A4). In \cite{chen2016tracking}, the authors addressed the target following with consideration of A1,A2,A4 and A5. The authors designed a hierarchical planner to consider the trade-off between optimality and online computation where a corridor is pre-planned to ensure safety and then a smooth path is generated to minimize the high-order derivatives in the following phase. Especially, \cite{chen2016tracking} performs prediction of target movement with a polynomial regression over past observations. Still, the prediction did not consider obstacles (A5) and the  occlusion of the target was not included in their multi-layer planner, having difficulty in handling A3.        

Regarding the target prediction in target following tasks, \cite{vsvec2014target} included the obstacles for the formulation of prediction directly tackling A5. The authors performed Monte-Carlo sampling to estimate the distribution of future target position. In the paper, however, the dynamics and set of possible inputs of target were assumed to be known as a prior, which is difficult to directly applied to general cinematic scenario. Also, the author restricted the homotopy of the solution path of the robot assuming discrete selection of actuation input. This method might not be able to achieve enough travel efficiency as pointed out in A1. 

To the best of our knowledge, there is only few research which  effectively handle A1 to A5 simultaneously for drones to be employed in the considered cinematic or chasing scenarios.
In this letter, we make the following contributions as extension of our previous work \cite{jeon2019online}
\begin{itemize}
    \item An integrated framework for motion strategy is proposed from prediction module to chasing planner, which could achieve desired performance mentioned A1-A5 in our cinematic problem setting. Especially, we validate the newly developed prediction module by examining its effectiveness for the proposed motion planner. 
    \item We validate our method by multiple challenging scenario and real world experiment. Especially, the tested real platform is implemented to operate fully onboard handling target detection, localization, motion planning and control.
\end{itemize}
 
The remainder of this paper is structured as follows: we  first describe problem proposition and overall approach. In the subsequent section, method for target prediction for a future time window is proposed in section \ref{section:predict}, which is followed by a hierarchical chasing planner design\; in section   \ref{section:preplanning} and \ref{section:smooth path}.

%\IEEEpubidadjcol

% An example of a floating figure using the graphicx package.
% Note that \label must occur AFTER (or within) \caption.
% For figures, \caption should occur after the \includegraphics.
% Note that IEEEtran v1.7 and later has special internal code that
% is designed to preserve the operation of \label within \caption
% even when the captionsoff option is in effect. However, because
% of issues like this, it may be the safest practice to put all your
% \label just after \caption rather than within \caption{}.
%
% Reminder: the "draftcls" or "draftclsnofoot", not "draft", class
% option should be used if it is desired that the figures are to be
% displayed while in draft mode.
%
\begin{figure}[!t]
\centering
\includegraphics[width=0.45\textwidth]{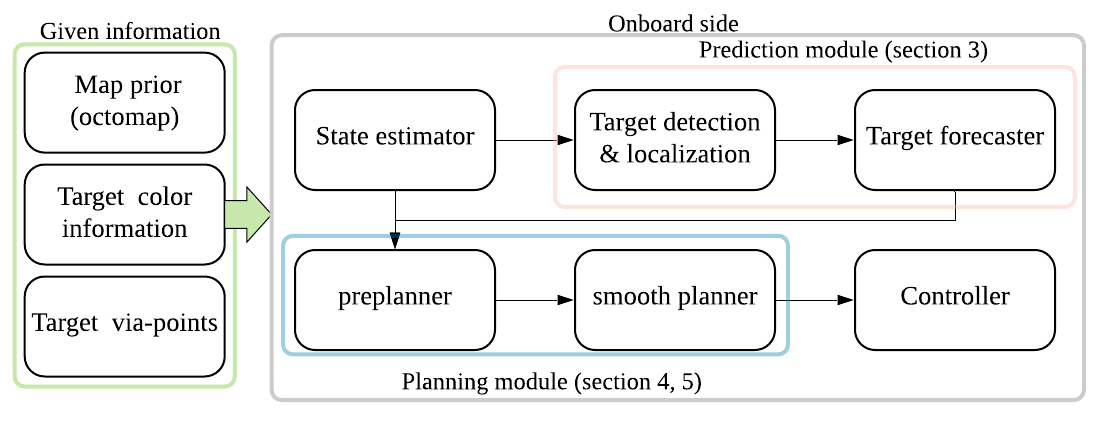}

\caption{A diagram for system architecture: we suppose the camera-drone has prior knowledge of the environment and target color for detection. Based on them, we implement a fully onboard system for automatically following a moving target with drone. For state estimator of drone, we utilize the ZED internal visual odometry and pixhawk is used for flight control. Target future motion prediction and chasing planner are proposed in this letter. }
\label{fig_system}
\end{figure}

\section{Overview}
\label{overview}
Here we outline and formalize the given information and the desired capabilities of the proposed method. We assume that the filming scene is available in the form of octomap before filming. It is also assumed that the actor will pass a set of via-points in sequence for a filming purpose. The viapoints are known a priori while the arrival time for each point is not available. As an additional specification on the behavior of target, it is assumed to move along a trajectory by minimizing high-order derivatives such as acceleration as assumed in \cite{chen2016tracking}. Additionally, we assume that the target object is visible from a drone at the start of the mission within the limited  field-of-view (FOV) of the drone.

Based on these settings, we focus on a chasing planner and target prediction module which can handle A1-A5 simultaneously as mentioned in section \ref{intro}. Additionally, the chasing strategy optimizes the total travel distance and the efforts  to maintain a desired relative distance between the drone and object.     

\section{Target future trajectory prediction}
\label{section:predict}

This section describes target future motion estimation utilizing the observation history and prior map information. Here the terms \textit{path} and \textit{trajectory} are differentiated for clarity as described in \cite{gasparetto2015path}. A path refers to a geometric path while trajectory is time-parameterized path. The proposed prediction module generates a geometric prediction path first, which will be followed by time estimation for the each point on the path.  

\begin{figure}[!t] 
\centering
\includegraphics[width=0.3\textwidth]{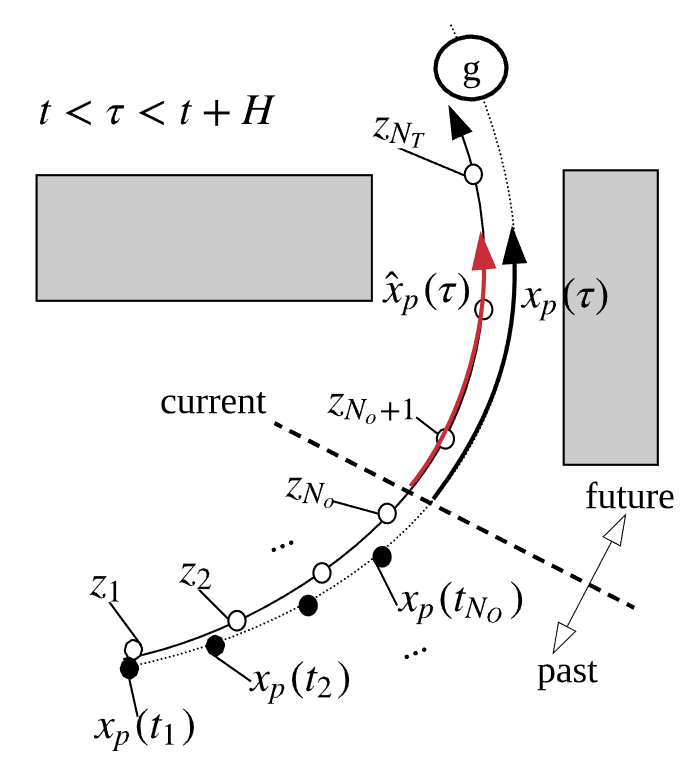}
\caption{Prediction of target motion over a horizon $(t,t+H]$.}
\label{fig_predict}
\end{figure}

\subsection{Target path prediction}
As mentioned in section \ref{overview}, we assume that the sequence of via-points of the target is available as $G=\{ g_1, g_2, ... , g_M\},g_i \in \mathbb{R}^3 $ which is supposed to be passed in order and the arrival time for each point is not preset. Also, let us denote the value of the Euclidean signed distance field (ESDF)  at a position $x\in \mathbb{R}^3$ of the prior map as $\phi(x)$. We denote the position of object at time $\tau$ as $x_p(\tau)\in \mathbb{R}^3$. 

Now, let us assume that the drone has gathered target observation at discrete time steps $t_1,t_2,..,t_{N_o}$ and write $x_p(t_n) \; (n=1,..,N_o)$ as $x_{p,n}$. Additionally, let us consider a situation where the target heads to $g\in G$ after passing the previous waypoint. For a prediction callback time $t>t_{N_o}$ and a future horizon $H$, we want to forecast the future target trajectory $x_p(\tau) \; (\tau \in (t,t+H])$ with the estimated trajectory $\hat{x}_p(\tau)$. To obtain $\hat{x}_p(\tau)$, a positional path $\xi = [z_1^T , z_2^T, ..., z_{N_T}^T]^T,\; z_i \in \mathbb{R}^3$ where $N_T > N_o$ is generated first to provide a geometric prediction until the point where the target reaches $g$ by solving the following optimization.         %%% HJK   
    \begin{equation}
    \label{eqn_predict_optim1}
    \begin{aligned}
             &   \underset{\xi}{min} &&  \underbrace{\dfrac{1}{2} \sum_{n=1}^{N_o}\mathrm{exp}(\gamma n) \| z_n-x_{p,n}\|^2}_{observation } + \\ & & & \underbrace{\dfrac{1}{2} \sum_{n=1}^{N_T-2} \| z_n-2 z_{n+1}+z_{n+2}\|^2}_{2nd\; derivatives }  + \dfrac{1}{\rho}\underbrace{\sum_{n=1}^{N_T}f_{obs}(z_n)}_{obstacle}  
    \end{aligned}
    \end{equation}
where $\gamma$ in the first term is a positive constant for weighting more to error of recent observation. The second term implies the assumption that the target will minimize its derivatives for the sake of its actuation efficiency. The function $f_{obs}$ in the last term is a non-convex cost function to reflect the safe behavior assumption of target (see  \cite{ratliff2009chomp} for more details of the functional), which is computed based on the $\phi(x)$ of the prior map information.
\cref{eqn_predict_optim1} can be arranged into the below, which is the standard form for covariant optimization \cite{ratliff2009chomp}. 

    \begin{equation}
    \label{eqn_predict_optim2}
    \begin{aligned}
             & &  \underset{\xi}{min} &&  \underbrace{\dfrac{1}{2}\rho \| A \xi -b\|^2}_{prior \; term}  + \underbrace{f_{obs}(\xi)}_{obstacle\; term}  
    \end{aligned}
    \end{equation}
(\ref{eqn_predict_optim2}) is solved with the following covariant update rule where $\alpha$ is a step size.
    \begin{equation}
    \label{eqn_predict_update}
        \Delta \xi = -\alpha (A^{T}A)^{-1} (\rho(A^{T}A\xi-A^{T}b)+\nabla f_{obs}(\xi))
    \end{equation}
    
From \eqref{eqn_predict_optim1} -- \eqref{eqn_predict_update}, a geometric path of target is predicted using $z_{N_o+1} ,..., z_{N_T}$ until $g$ (see fig. \ref{fig_predict}). In the following step, the path is endowed with time to complete prediction.   

\subsection{Time prediction} In this subsection, we will allocate time knots $t_1 , t_2, ... , t_{N_o}, t_{N_o+1},..,t_{N_T}$ for each point in $\xi$ with the constant velocity assumption for the simplicity.
For the points $z_1,z_2,...,z_{N_o}$ which was used to regress on the past history of target, we simply assign the observation time stamps $t_1,..., t_{N_o}$. For the predicted position $z_n$ ($n > N_o$), the following recursion is used for allocating times. 

\begin{equation}
\begin{aligned}
    t_{N_o+1} = t_{N_o} + \dfrac{\|z_{N_o+1} -z_{N_o}\|}{v_{avg}} \\
    t_{N_o+2} = t_{N_o+1} + \dfrac{\|z_{N_o+2} -z_{N_o+1}\|}{v_{avg}}\\ \label{eqn_time_alloc}
    \dots \\
    t_{N_T} = t_{N_T-1} + \dfrac{\|z_{N_T} -z_{N_T-1}\|}{v_{avg}}
\end{aligned}
\end{equation}
where\;
$ v_{avg} = \dfrac{\sum_{n=1}^{N_o-1}\|x_{p,n} - x_{p,n+1}\|}{t_{N_o} - t_{1}}$ 
represents the average speed during the collected observation. The passing times for the points obtained in \cref{eqn_predict_optim1} are estimated with the constant velocity assumption based on $v_{avg}$. With this allocation, the future trajectory of target for a time window $(t,t+H]$ is predicted with the following interpolation: 

\begin{equation}
    \hat{x}_p(\tau) = \dfrac{(t_{n+1} - \tau)z_n + (\tau-t_n)z_{n+1}}{t_{n+1} - t_n} \;\; (t_n<\tau<t_{n+1})    
\end{equation}

In our videography setting which will be introduced in Sec. \ref{section:result}, single prediction optimization routine runs at 30-50 Hz showing its real-time performance. This helps us to re-trigger prediction when the estimation error exceeds a threshold on-the-fly.

\section{Preplanning for chasing corridor}
\label{section:preplanning}
This section introduces a method for generating a chasing corridor in order to provide the boundary region for the chasing drone's trajectory. We first explain a metric to encode safety and visibility of the chaser's position, which is utilized as objectives in computing the corridor. Then, the corridor generation by means of graph-search is described. Several notations are defined adding to $x_p$ as follows: 
\begin{itemize}
    \item[--] $v_{c}\in\R$ : Position of a chaser\;(drone).
    \item[--] $v_{p}\in\R$ : Position of a target.
    \item[--] $L(v_{1},v_2) = \{v|\; sv_1+(1-s)v_2,\  0 \leq t \leq 1\}$ : The line segment connecting $v_1, v_2 \in \R$. 

    \item[--] $\mathbf{\chi}\subset\R $ : Configuration space.
    \item[--] $\mathbf{\chi}_{free} = \{{v}|\;P({v})<\epsilon\}$ : Free space in $\configspace$, i.e. the set of points where the probability of occupancy $P({v})$ obtained from octomap is small enough. 
    \item[--] $\mathbf{\chi}_{obs} = \configspace \setminus \free$ : Space occupied by obstacles. 

    \item[--] $\mathbf{\chi}_{vis}(v_{p}) = \{v|\; L(v,v_{p}) \cap \obs = \emptyset \} $ : A set of visible vantage points for a target position $v_p$.  
    \item[--] $\mathbf{\chi}_{occ}(v_{p}) = \configspace \setminus \vis $ : A set of occluded vantage points for a target position $v_p$. 

\end{itemize}

\begin{figure}[!t] 
\centering
\includegraphics[width=0.34\textwidth]{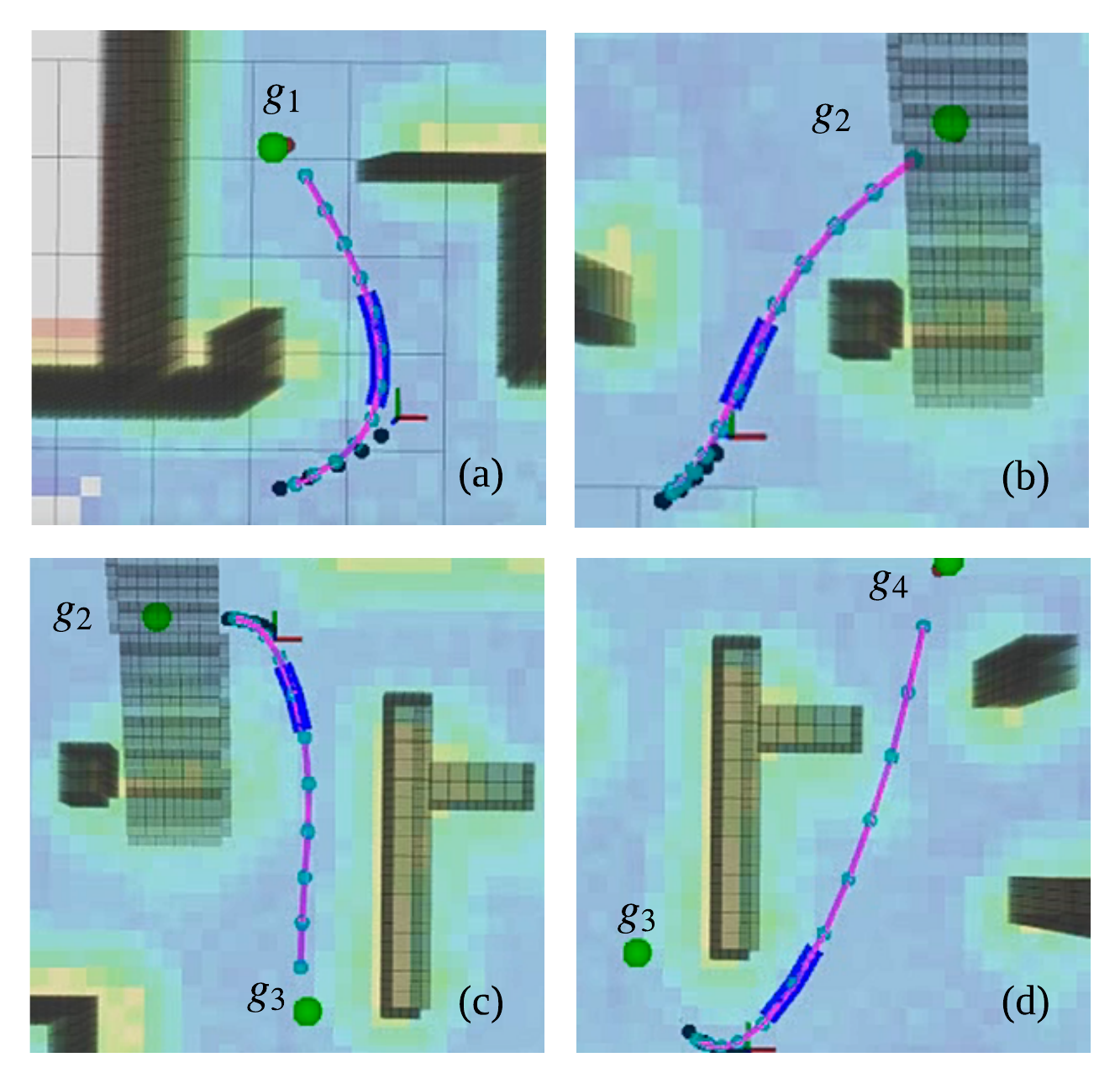}
\caption{Prediction result in the \textit{complex city} cinematic scenario (the detailed settings can be found in \ref{section:result}). In the figure, black dots denote past observation for actor in the buffer and green dot for target via-points sequence $G$ . Blue points and Magenta line denote geometric path $\xi$ while thick blue line means trajectory estimation over horizon $H$. Also, the cost function $f_{obs}$ based on ESDF $\phi(x)$ is illustrated in jet-colormap.}
\label{fig_predict_result}
\end{figure}
 
\subsection{Metric for safety and visibility}
\begin{figure}[t] 
\centering
\includegraphics[width=0.4\textwidth]{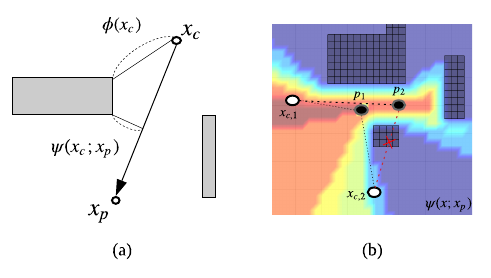}
\caption{(a): safety metric $\phi(x_c)$ and visibility metric $\psi(x_c;x_p)$ for a target position $x_p$ and drone $x_c$. (b): Visibility field for $x_p$ in colormap.  Red denotes higher visibility and the occluded region $\mathbf{\chi}_{occ}$ is illustrated with the same uniform  color (dark blue) for simplicity. As an illustrated example, we consider the case where the object moves $p_1 \to p_2$ for a short time. Both positions $x_{c,1}$ and $x_c{c,2}$ are able to see $p_1$.  While the camera-drone at $x_{c,1} $ can still observe the target at $p_2$, it fails to maintain the visibility at $x_{c,2}$.} 
\label{fig_metric}
\end{figure}

For the safe flight of the camera-drone $x_c$, we reuse ESDF $\phi(x)$ as it can measure the risk of collision with nearby obstacles. Here, $\phi(x_c)$ is used as a constraint in graph construction so that drone can maintain a safe clearance during entire planning horizon. Now, the visibility metric is introduced so that we can encode how robustly the drone can maintain its sight against occluding obstacles and unexpected motion of the target in the near future. For a target position $x_p$ when seen from a chaser position $x_c$ with line of sight (LOS) $L(x_p,x_c)$, we define the below as visibility:

\begin{equation}
\label{eqn_visfield}
    \psi(x_c;x_p) = \underset{L(x_c,x_p)}{\text{min}} \phi(x)
\end{equation}
In the actual implementation, (\ref{eqn_visfield}) is calculated over the grid field. That is, we can evaluate (\ref{eqn_visfield}) with a simple min operation while iterating through  voxels along $L(x,x_p)$ with the linear time complexity. Because (\ref{eqn_visfield}) means the minimum distance between obstacle and LOS connecting the object and drone, its small value implies that the target could be lost more easily than the higher value as illustrated in fig. \ref{fig_metric} - (b). The proposed metric possesses multiple advantages. First, it can be directly computed from reusing ESDF which was utilized for the target prediction and safety constraint for drone, without further complex calculation. Second, it can be defined without restriction of shape of obstacle in contrast to the research such as \cite{nageli2017real},  \cite{penin2018vision} and \cite{bonatti2018autonomous}. Detailed explanation on the advantage and properties of the proposed metric is referred to our previous work \cite{jeon2019online}.     

\subsection{Corridor generation}
Based on the proposed metric for safety and visibility, computation of the sequence of corridors for chasing is explained here. Before that, we plan a sequence of viewpoints for a time horizon $(t,t+H]$ as a skeleton for it. Let us assume that the drone is at $x_c(t)$ and target prediction $\hat{x}_p(\tau) \; (t<\tau \leq t+H)$. For a window $[t,t+H]$, time is discretized as $t_0,t_1,t_2,...,t_N$ and we rewrite $\hat{x}_p(t_k) = \hat{x}_{p,k}$, $x_c(t_k) = x_{c,k}$. Here, the sequence of viewpoints  $\sigma = \{v_k\}_{1\leq k \leq N}$ is generated where point $v_k\in \mathbb{R}^3$ is selected from a set $V_k \subset \{x_i| d_{l} \leq || x_i-\hat{x}_{p,k}|| \leq d_{u}, \; x_i \in \mathbf{\chi}_{vis}(\hat{x}_{p,k})\}$. $x_i$ denotes a discrete point in a given grid. $d_{l}$ and $d_{u}$ are the minimum and maximum distance of tracking. The discrete path $\sigma$ is obtained from the following discrete optimization.

\begin{equation}
\label{eqn: optim for preplanning}
\begin{aligned}
& \underset{{\sigma}}{\text{argmin}}
&& \sum_{k=1}^{N}c(v_{k-1},v_k) \\
& \text{subject to} & &\text{${v}_{0} = x_{c,0}$} \\
& & &\underset{x \in L(v_{k-1},v_{k})}{\text{min}}\phi(x) \geq r_{safe} \\
& & & v_k \in  V_k  \\
& & & \text{$\lVert {v}_{k-1} -{v}_{k}\rVert \leq d_{max}$} \\ 
& \mathrm{where} & &c(v_{k-1},v_k)=\underbrace{ \lVert {v}_{k-1} -{v}_{k}\rVert^2}_{\mathrm{interval \:distance}} \: + \underbrace{w_{v} \; {c_{v}}({v}_{k-1},{v}_{k})}_{\mathrm{visibility}}  \\ & & & + w_d \underbrace{(\lVert {\hat{x}}_{p,k} -{v}_{k}\rVert - d_{des})^{2}}_{\mathrm{tracking\: distance}} \\
\end{aligned}
\end{equation}

\begin{figure}[t] 
\centering
\includegraphics[width=0.48\textwidth]{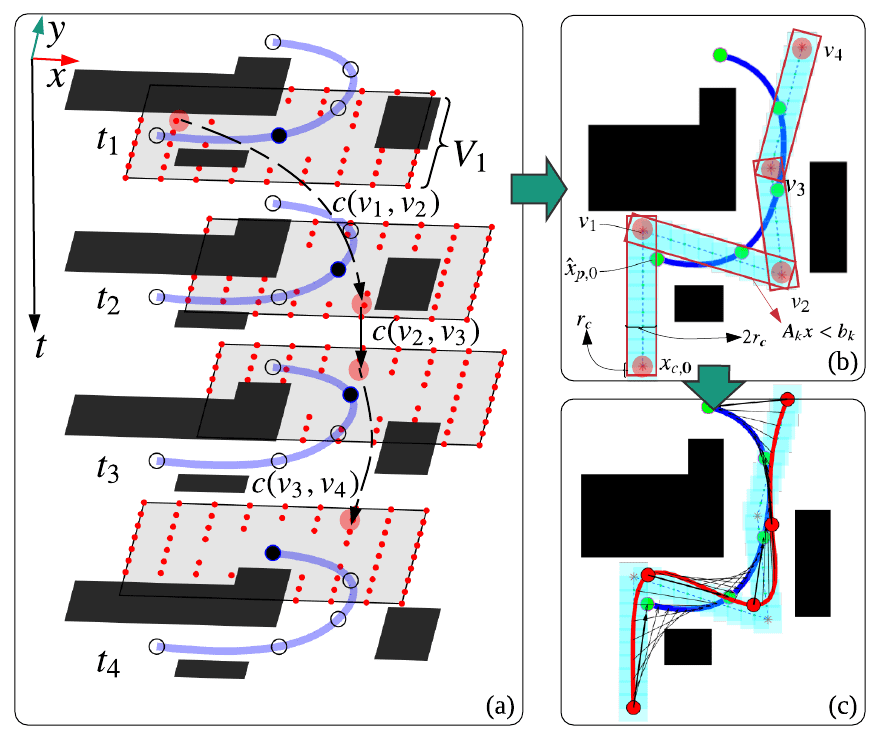}
\caption{(a): Illustrative example for graph construction. The blue line denotes target prediction for a time window. Red dots denote the elements in $V_k$ at each time step. We connect two nodes of two consecutive sets $V_k,\;V_{k+1}$ as a directional edge if the second and fourth constraints in (\ref{eqn: optim for preplanning}) hold. {Top}: chasing corridor based on the skeleton $\sigma$. The red edged boxes denote corridors between $v_k$ and $v_{k+1}$ with width $r_c$. (c): smooth path (red) is generated within the corridors from (b). LOS stamps are also drawn with black arrows.} 
\label{fig_corridor}
\end{figure}

The objective function in (\ref{eqn: optim for preplanning}) penalizes the interval distance between each point and rewards the high score of visibility along path. The second term is defined by
\begin{equation}
\begin{aligned}
& c_v(v_{k-1},v_{k})= \\&\Bigg(\sqrt{\int_{L(v_{k-1},v_k)} \psi(x;\hat{x}_{p,k-1})dx \int_{L(v_{k-1},v_k)} \psi(x;\hat{x}_{p,k})dx}\Bigg)^{-1}.
\end{aligned}
\end{equation}
The last term in \eqref{eqn: optim for preplanning} aims to keep the relative distance between drone and object as $d_{des}$. $w_v$ is the weight for visibility and $w_d$ is for relative distance. Among the constraints, the second one enforces a safe clearance $r_{safe}$ of each line $L(v_{k-1},v_k)$ and the third constraint means that $v_k$ should be a visible viewpoint for  the predicted target at $\hat{x}_{p,k}$. The last constraint bounds the maximally connectable distance between two points $v_k$, $v_{k+1}$ in subsequent steps. More details on the method for building a directional graph to solve the above discrete optimization is explained in our previous research\cite{jeon2019online} and fig. \ref{fig_corridor}-(a). 

From $\sigma$ computed from (\ref{eqn: optim for preplanning}), we generate a set of corridors connecting two consecutive viewpoints in $\sigma$ as visualized in Fig \ref{fig_corridor}-(b). Once the width of corridor $r_c < r_{safe}$ is chosen, we can write the  box region connecting $v_{k-1}$ and $v_{k}$ as a linear inequality $A_k x \leq b_k$. The corridor $A_k x \leq b_k$ is depicted with red rectangles in fig. \ref{fig_corridor}-(b). Due to the formulation of (\ref{eqn: optim for preplanning}), every point in $A_k x \leq b_k$ maintains a safe margin $r_c < r_{safe}$ and from the viewpoints $v_k$  the predicted point $\hat{x}_{p,k}$ can be observed without occlusion by obstacles. Also, for a large value of $w_v$ and small enough $d_{max}$, we empirically found that every point in corridor $A_k x \leq b_k$ can maintain visibility for the prediction $\hat{x}_p(\tau)\;$ for $\tau \in [t_{k-1},t_k]$.

\section{Smooth path generation}
\label{section:smooth path}

In the previous section, the procedure to select viewpoints $v_k$ and corridor $A_kx\leq b_k$ was proposed, which  was computed by optimally considering visibility and travel distance while ensuring safety. In this section, we generate a dynamically feasible trajectory for position and yaw $[x_c(\tau)^T,\; r(\tau)]^T\in\mathbb{R}^4$ using $v_k$ and $A_kx\leq b_k$. The position trajectory $x_c(\tau)$ is represented with piece-wise polynomials as below:
\begin{equation}
x_c(\tau) =  
\begin{cases}
 \sum_{i=0}^{K}p_{1,i}\tau^i & (t_{0} \leq \tau < t_{1}) \\
 \sum_{i=0}^{K}p_{2,i}\tau^k & (t_{1} \leq \tau < t_{2}) \\
...&\\
 \sum_{i=0}^{K}p_{N,i}\tau^i & (t_{N-1} \leq \tau < t_{N}) \\
\end{cases}
\end{equation}
Where $p_{k,i} \in \mathbb{R}^3\; (k=1,...,N,\; i=1,...,K)$ is coefficient and $K$ denotes the order of the polynomial. 
Polynomial coefficients of the chaser's trajectory are computed from the  optimization below \eqref{eqn:cont opti}. The planning of yaw $y(\tau)$ was done so that $x_c(\tau)$ heads toward ${x}_p(\tau)$ at each time step if observation of the target at $\tau$ is acquired.   

\begin{equation}
\label{eqn:cont opti}
\begin{aligned}
& \underset{}{\text{min}}
&& \int_{t_{0}}^{t_{N}} {\lVert{{x_c}^{(3)}(\tau)}\rVert}^2 d\tau \:+\: \lambda \sum_{k=1}^{N}{\lVert{x_c(t_{k})}-v_{k}\rVert}^2 \\ 
& \text{subject to} & &  x_c(t_{0}) = x_{c,0}  \\
& & &   \dot{x}_c(t_{0}) = \dot{x}_{0}\\
& & &   \ddot{x}_c(t_{0}) = \ddot{x}_{0}\\
& & &   A_k x_c(\tau) \leq b_k \; (t_{k-1}<\tau <t_{k},\; k=1,...,N)  
\end{aligned}
\end{equation}
 Our optimization setup tries to minimize the magnitude of jerk along the trajectory and the deviation of $x_c(t_k)$ from viewpoints $v_k$ where $\lambda$ is an importance weight. In the constraints, $x_0,\;\dot{x}_0$ and $\ddot{x}_0$ is the state of drone when the planning was triggered and used as the initial condition of the optimization. Additionally, we enforce continuity conditions on the knots. The last constraint acts as a box constraint so that the smooth path is generated within the chasing corridors for the purpose of safety and visibility. As investigated in \cite{mellinger2011minimum}, $x_c(\tau)$ can be executed by the virtue of differential flatness of quadrotors. \eqref{eqn:cont opti} can be solved efficiently with the algorithm such interior point  \cite{mehrotra1992implementation}. The overall algorithm is summarized in \textbf{Algorithm 1}. During mission, we predict target future trajectory for a time window $H$ with several recent observation by solving \eqref{eqn_predict_optim1}. If observation becomes unreliable where the accumulated estimation error exceeds a defined threshold, prediction is re-triggered. Based on observation, the chasing planner yields a desired trajectory for the chaser by pre-planning and generating a smooth path to be executed during a corresponding horizon. This loop continues until the end of the videographic mission.

\begin{algorithm}
\DontPrintSemicolon
\SetAlgoLined
\SetKwInOut{Input}{Input}\SetKwInOut{Output}{Output}
\SetKwInOut{given}{Input}
\SetKwInOut{param}{Parameter}
\SetKwInOut{initialize}{Initialize}
\SetKwFunction{find}{find}
\SetKwFunction{append}{append}
\SetKwFunction{predict}{Predict}
\SetKwFunction{planning}{Planning}
\given{SDF $\phi(x)$,\; target via-points $g_1 \to g_2 \to ,...,\to g_N$,\; receding horizon window $H$,\; time discretization $N$}
\initialize{$g=g_1,\; \texttt{accumErr}=0$}
\For{$t=t_s$  \KwTo $t_f$}{ // \textit{from mission start to finish} \\
    \ForAll{$v_{n-1} \in V_{n-1}$}{
        \texttt{observ}.\append{$x_p(t)$} \\
        \If{$\texttt{accumErr}>\epsilon$}{
            $\hat{x}_p(\tau)$ = \predict{\texttt{observ}, $g$}\\ 
            ${x}_c(\tau)$ = \planning{$\hat{x}_{p,1}, \hat{x}_{p,2},..,\hat{x}_{p,N}$, $x_0,\dot{x}_0,\ddot{x}_0$}\\ 
            // $(t< \tau \leq t+H)$
        }
        $\texttt{accumErr}+=\lVert \hat{x}_p(t) -x_p(t) \rVert^2$ \\ 
        // accumulate estimation error \\    
        \If{$\lVert \hat{x}_p(t) -x_p(t) \rVert < \delta$}{
            $g$ = $g$.next 
            // if target is observed to reach $g$, update
        }
    }
}
\caption{Receding horizon chasing planner}
\end{algorithm}

\section{RESULTS}
\label{section:result}
% \begin{itemize}

%     % \item prediction performance comparison with other methods
%     % \item effect of prediction on chasing compared to the case where prior path of target available
%     \item computation speed
%     \item effect of high visibility weight on the occlusion duration under unreliable prediction situation
% \end{itemize}
\subsection{Simulations}
\begin{figure*}[t]
 \centering
  \includegraphics[width=0.9\textwidth]{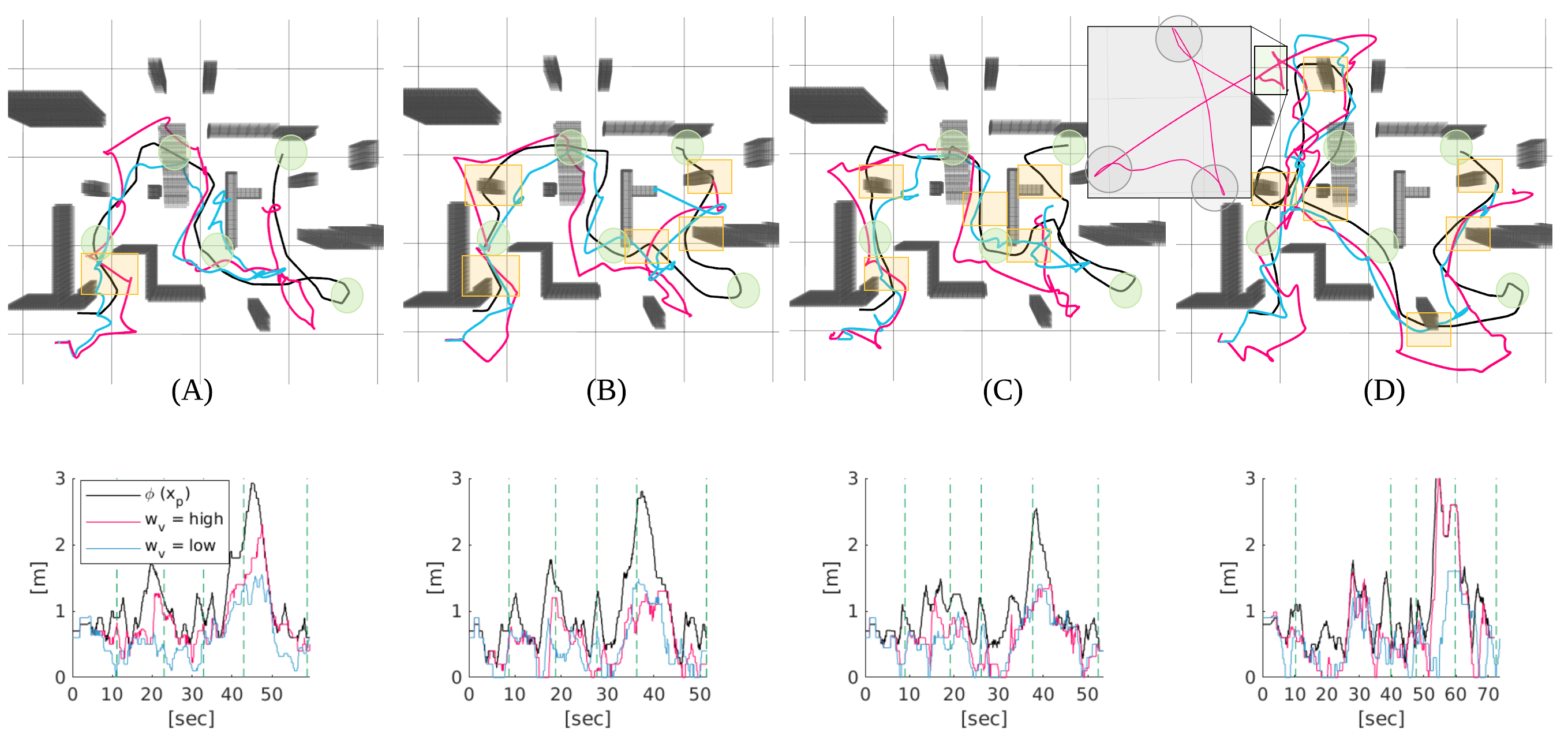}
  \caption{flight result for the four different trajectories of the target. \textbf{Top:} the target's history is denoted as a black line and the chaser's flight histories are depicted with skyblue for low visibility, and magenta for high visibility respectively. The size of the grid in the figure is 4[m]. From (A) to (D), the target moves faster and it is more unpredictable due to its hiding behind obstacles, which increases difficulty.     For all the  simulations, the target passes through five via-points one-by-one (green circles). The orange boxes denote the locations where the target was intentionally operated to hide behind obstacles with an abrupt maneuver. To confirm the smoothness of the flight trajectory, a zoom-in image is provided in (D). \textbf{Bottom:} history of distance field value of the target and visibility score for a small $w_v$ ($= 1.0$) and a large $w_v$ ($= 5.0$) are plotted. The dotted vertical line (green) denotes the the target's arrival time to each via-point.}
\label{fig_simulation}
\end{figure*}

\begin{figure}[h] 
\centering
\includegraphics[width=0.3\textwidth]{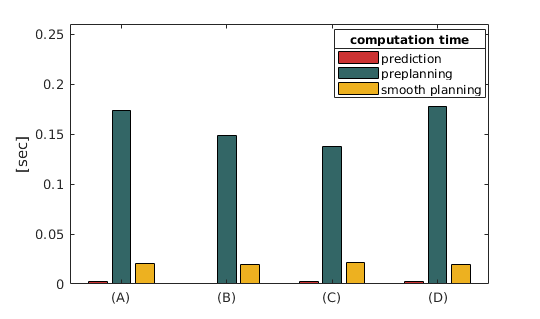}
\caption{Average computation time in each simulation scenario for three phases of prediction, preplanning and smooth planning.} 
\label{fig:computation_time}
\end{figure}

We validated the proposed algorithm in a dense environment with multiple target trajectories. For simulation, we used \textit{complex city} (see our previous work \cite{jeon2019online} for the 3D models) where five target via-points are defined as green circles as in \cref{fig_simulation}. \textit{Complex city} includes multiple non-convex obstacles, and the target was operated to hide behind the obstacles at the moment denoted as orange boxes (see \cref{fig_simulation}). In the simulation, we used \textit{rotors simulator} \cite{furrer2016rotors} for the chaser drone,  and the target (turtlebot) was manually operated with a keyboard.  A vision sensor is fixed on the drone (13\degree pitching down). Due to this, the elevation of LOS was limited when selecting the elements $v_k$ from $V_k$. All the virtual platforms operated in gazebo environment and the simulations were performed in Intel i7 CPU and 16GB RAM laptop. Boost Graph Library (GPL) was used for preplanning while qpOASES \cite{ferreau2014qpoases} was used to solve quadratic programming for the smooth planning phase. For the four target trajectories, chasing strategies with two different levels of visibility weights $w_v$ are tested, totalling 8 simulations. Other than the visibility weight $w_v$, the other parameters were set at the same value for all tests. For the simulation, we directly fed the current position of the turtlebot to the drone. The results are summarized in \cref{fig_simulation} (A)-(D) and \cref{table:result}. For each target scenario, the history of $\phi(x_p)$ is plotted in the bottom row in \cref{fig_simulation} where a small value of $\phi(x_p)$ implies difficulty for securing visibility due to the proximity of the target to obstacles. Planner for high visibility with $w_v=5.0$ tries to secure more visibility score $\psi(x_c;x_p)$ compared to planner with $w_v=1.0$. Specifically, the value of $\psi(x_c;x_p)$ of the  planner with $w_v=5.0$ was on average 24\% higher than the case with $w_v=1.0$. Also, the duration of occlusion was 42\% lower with $w_v=5.0$ in the four target trajectory cases. In contrast, the planner for low visibility with $w_v=1.0$ decreased the travel distance to 34\% on average compared to $w_v=5.0$. In all simulations, the safety of drone chaser was strictly satisfied during entire mission. The average computation times are summarized in \cref{fig:computation_time}. We observed that the entire pipeline of the receding horizon planner ran at 5-6Hz, showing the capability to re-plan fast enough in response to the unexpected target motion.      

\begin{figure}[h] 
\centering
\includegraphics[width=0.25\textwidth]{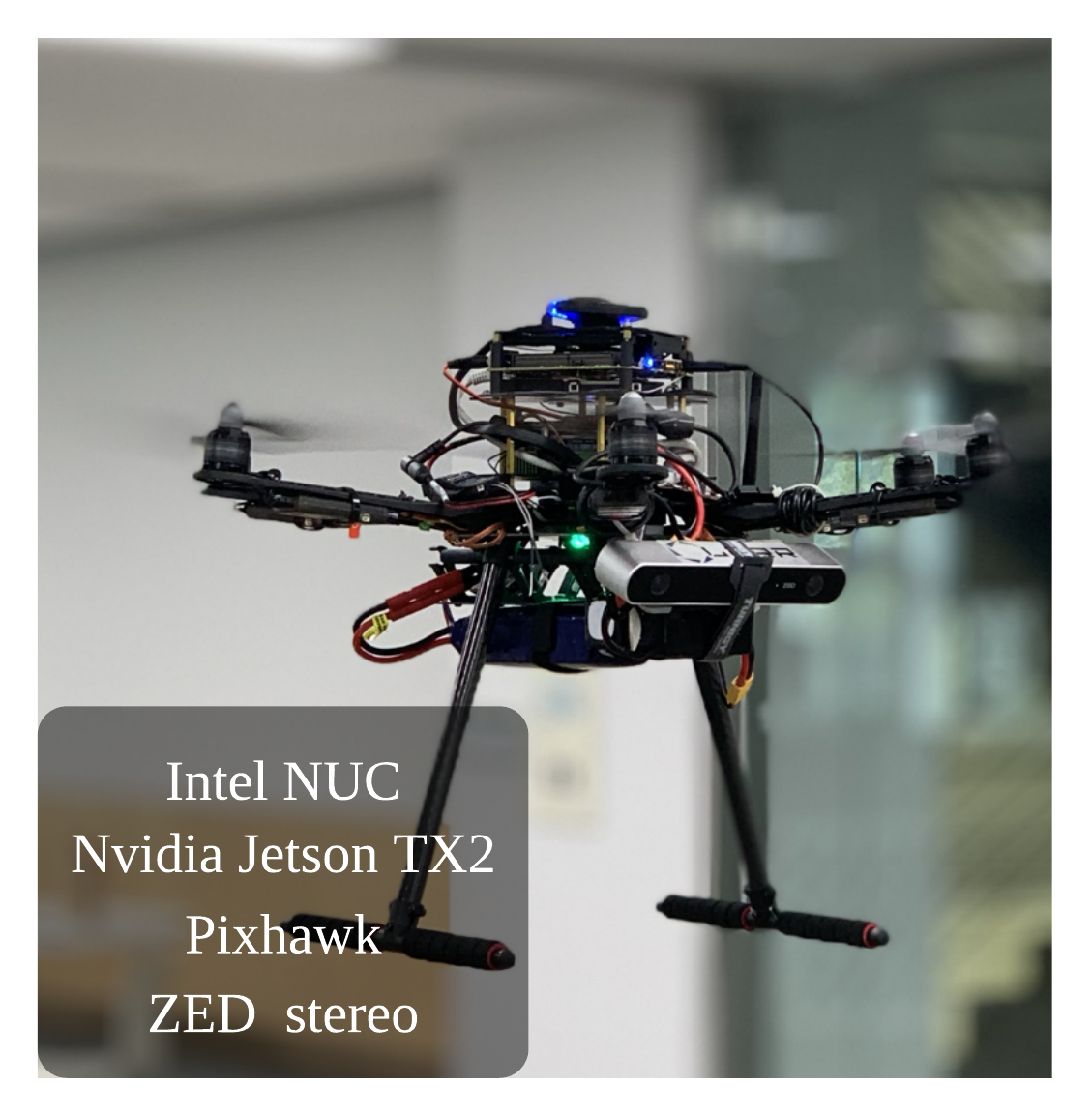}
\caption{The camera-drone for onboard implementation: for algorithm execution, intel NUC is used as core, and camera- and vision-related tasks such as visual odometry and target localization run on Jetson TX2. Pixhawk is employed as the flight controller.} 
\label{fig_hardware}
\end{figure}

\setlength\doublerulesep{0.4pt} 
\begin{table}[]
\centering
 \begin{adjustbox}{max width=0.48\textwidth}

\begin{tabular}{lcccccccc}
\toprule[1pt]
\hline
                      & \multicolumn{2}{c}{A} & \multicolumn{2}{c}{B} & \multicolumn{2}{c}{C} & \multicolumn{2}{c}{D} \\ \hline
                      \midrule[0.05pt] 
\multicolumn{1}{c|}{target speed [m/s]} & \multicolumn{2}{c}{0.36}  & \multicolumn{2}{c}{0.57}  & \multicolumn{2}{c}{0.67}  & \multicolumn{2}{c}{0.75}  \\

% \multicolumn{1}{c|}{pred. RMSE} & \multicolumn{2}{c}{0.1532}  & \multicolumn{2}{c}{0.1957}  & \multicolumn{2}{c}{0.2397}  & \multicolumn{2}{c}{0.2360}  \\ 
\hline
\multicolumn{1}{c|}{$w_v$} &   1.0        & 5.0       & 1.0          &  5.0         & 1.0          &    5.0       & 1.0          &  5.0         \\ \hline
\multicolumn{1}{c|}{avg. $\psi(x_c;x_p)$[m]} &     0.6084      &0.8433           &  0.4817          & 0.5330           & 0.5543           &          0.6051 &   0.5566        & 0.7791   \\

\multicolumn{1}{c|}{occ. duration [sec]} &    0.099       & 0          &        7.59   &    4.323       &   4.29        &       2.145    &      9.768     & 6.435          
\\

\multicolumn{1}{c|}{flight dist.[m]} &    40.7761       &   49.9106        &    34.8218       &     47.5424      &  36.3040         &   50.6000        &    55.7393       & 77.2377         
\\

\bottomrule[0.3pt]

\end{tabular}
\end{adjustbox}
\caption{Simulation result}
\label{table:result}
\end{table}

\setlength\doublerulesep{0.4pt} 
\begin{table}[h]
\begin{center}

% table for parameters
 \begin{adjustbox}{max width=0.48\textwidth}

\begin{tabular}{c|c c}
\toprule[1pt]\midrule[0.3pt]
\multicolumn{3}{c}{Parameters} \\
\hline
\midrule[0.3pt]
Type & Name & Value\\ \hline
\multirow{1}{*}{Common} & time window [s]  & $H=4$ \\ \hline
\multirow{4}{*}{Prediction} & obsrv. temporal weight  & $\gamma=0.1$ \\
& weight on prior term & $\rho=0.2$ \\
 & obsrv. pnts./pred. pnts & $N_o=4$ / $N_T=7$ \\ 
 & pred. accum. err. tol.[m]  & 1.0  \\ 

 \hline

\multirow{7}{*}{Preplanning } & tracking distance weight & $w_d = 5.5 $ \\
& desired tracking dist.[m] & $d_{des} = 2.5 $\\ 
& maximum connection[m] & $d_{max} = 2.0 $\\ 
& lower and upper bounds of relative dist. [m] & $d_l=1.0$ / $d_u=4.0$ \\  
& resolution[m] & $0.4$\\
& time step & $N = 4$\\
& safe margin[m] & $r_{safe} = 0.3$\\
 \hline

\multirow{2}{*}{Smooth planning}
 & waypoint weight& $\lambda = 0.5$ \\
 & polynomial order & $ K = 6$ \\
 
& safe tol.[m] & $r_{c} = 0.2 $\\
& tracking elev. $\theta_r$ & $ 20^{\circ}\leq \theta_r \leq 70^{\circ} $\\
\bottomrule[0.3pt]
\bottomrule[0.3pt]
\end{tabular}
\label{Table:params}
\end{adjustbox}
\end{center}
 \caption{Common parameters for simulations}
\end{table}

\subsection{Real world experiment}
\begin{figure*}[t] 
\centering
\includegraphics[width=0.9\textwidth]{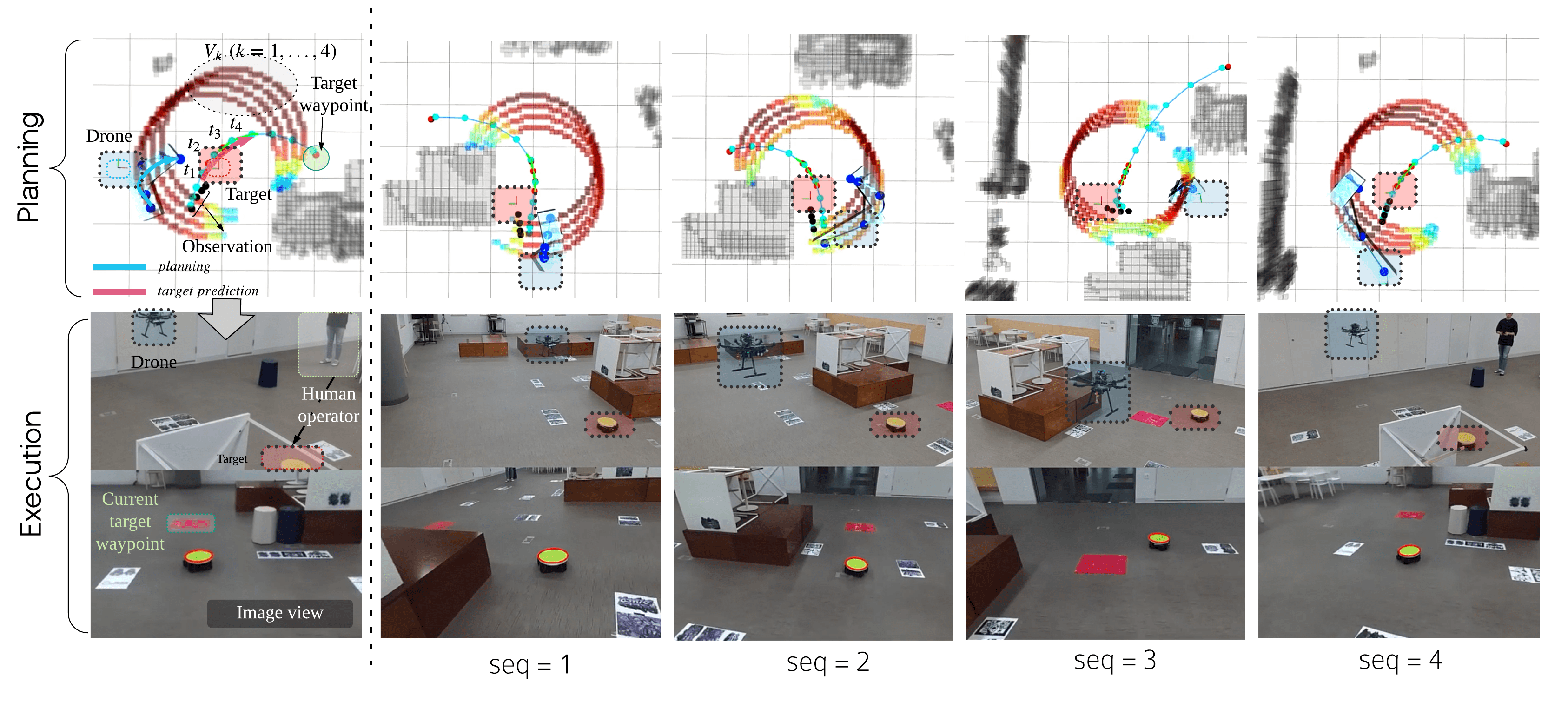}
\caption{\textbf{Top:} illustration of the chasing algorithm and the experiment. The localized target and drone are denoted as red and blue dotted boxes respectively. The target's future trajectory for 4 seconds is predicted as shown in the red line. The elements of $V_k\; (k=1,..,4)$ for each time step are depicted with their visibility score (red color denotes higher score). The drone plans its chasing path over a short horizon based on the preplanned points (dark blue sphere). \textbf{Bottom}: the snapshots of the drone flight for chasing a target, and camera view of the drone. The target is detected with an enclosed ellipse in the view. The entire path taken by target and drone is visualized in \cref{fig_intro}.} 
\label{fig_exp}
\end{figure*}

We tested the proposed method in an actual experiment in an indoor  classroom without GPS. In the test, the drone is equipped with ZED (stereo vision sensor) and pixhawk2 auto-pilot for  flight control. For visual odometry (VO), we used ZEDfu (the internal VO algorithm of ZED). The vision-related algorithm ran on Jetson TX2 while planning and control was processed in the onboard computer (NUC7i7BNH) (see \cref{fig_hardware}). The target is turtlebot waffle PI and a disk with green color was attached on top of the turtlebot to simplify the detection from the drone. The target was operated manually by a human operator with linear velocity of 0.2-0.3 m/s. In order to obtain the position of the target, we thresholded HSV (hue, saturation, value) color to segment the target into an ellipsoid. We extracted the pointcloud of the center of the ellipsoid, to which we applied a smoothing filter to finalize the target's position. The experimental environment and the path taken by the target and chaser are shown in fig. \ref{fig_intro} with stamps of bearing vector. The whole pipeline runs at 10 Hz in our parameter settings for the experiment. In the experiment, we set $d_{des}1.3=$m, $d_{l}=1.0$m  and $d_{u}=1.5$m. The grid size used for octomap and the computation of visibility score is 0.1m. $w_v=5.0$ was set to the visibility weight. The entire path history for the drone and the target is plotted in \cref{fig_intro}. The result of each planning trigger can be found in \cref{fig_exp}.

\section{Conclusion and future works}
In this letter, we proposed a chasing planner to handle safety and occlusion against obstacles. The preplanning phase provides a chasing corridor where the objectives such as visibility, safety and travel distance are optimally incorporated. In the smooth planing, a dynamically feasible path is generated based on the corridor. We also proposed a prediction module which allows the camera-drone to forecast the future motion during a time horizon, which can be applied in obstacle cases. The whole pipeline was validated in various simulation scenario, and we implemented real drone which operates fully onboard to perform autonomous videography. We also explored the effect of visibility weights to the two conflicting objectives: travel distance and visibility. From the validations, we found that the chaser was able to handle multiple hiding behavior of target effectively by optimizing the visibility. In the future, we will extend the proposed algorithm for the multi-target chasing scenario. Also, we plan to enhance the algorithm for the case of unknown map where the drone has to explore to gather information to generate more efficient  trajectory.

% or




%\bibliographystyle{IEEEtran}
\bibliographystyle{ieeetr}

\bibliography{bibliography}



% \begin{IEEEbiography}[{\includegraphics[width=1in,height=1.25in,clip,keepaspectratio]{profile_Jeon}}]{Boseong Felipe Jeon}
% \end{IEEEbiography}

% \begin{IEEEbiography}[{\includegraphics[width=1in,height=1.25in,clip,keepaspectratio]{profile_prof}}]{H. Jin Kim}
% \end{IEEEbiography}

%\newpage


%\vfill


\end{document}